\documentclass{article}
\usepackage{spconf,amsmath,graphicx,hyperref}

\usepackage{url}
\usepackage[T1]{fontenc}
\usepackage{multirow}
\usepackage{booktabs}

\begin{document}

\title{Preservation of Language Understanding Capabilities in Speech-aware Large Language Models}


\name{
  \begin{tabular}{c}
  Marek Kubis$^{\star}$ \qquad Paweł Skórzewski$^{\star}$ \qquad Iwona Christop$^{\star}$  \qquad
  Mateusz Czyżnikiewicz$^{\dagger}$ \\ Jakub Kubiak$^{\dagger}$ \qquad Łukasz Bondaruk$^{\dagger}$ \qquad Marcin Lewandowski$^{\dagger}$
\end{tabular}
}

  \address{$^{\star}$Adam Mickiewicz University,
   ul. Uniwersytetu Poznańskiego 4, 61-614 Poznan, Poland\\
  $^{\dagger}$
   Samsung R\&D Institute Poland, Plac Europejski 1, 00-844 Warszawa, Poland}

\maketitle

\begin{abstract}
The paper presents C3T (Cross-modal Capabilities Conservation Test),
a new benchmark for assessing the performance of speech-aware large
language models. The benchmark utilizes textual tasks and a voice cloning text-to-speech
model to quantify the extent to which language understanding
capabilities are preserved when the model is accessed via speech input.
C3T quantifies the fairness of the model for different categories of
speakers and its robustness across text and speech modalities.
\end{abstract}
\section{Introduction}

The rapid advances in research and development of large language models have led to the creation of
new, more demanding benchmarks that address a wide variety of properties.
Ranging from factual knowledge \cite{hendrycks2021mmlu}, through logical reasoning
\cite{clark2018arc} and mathematical tasks \cite{cobbe2021gsm8k}, to the theory of
mind \cite{kosinski2024tom}, the capabilities of large language models are studied
under increasingly demanding conditions.
However, with the advent of multimodal models, there is a need to develop evaluation
procedures that go beyond the textual domain.

In this paper, we focus on evaluating the models that take speech as input.
One of the basic problems that arises when transitioning from a textual LLM to a
speech-aware LLM architecture is checking whether the language understanding capabilities of the model are
preserved.\footnote{We denote by \emph{textual LLM} models that provide only textual input. By
  \emph{speech-aware LLM} we mean any model that provides at least a textual and a speech interface.}
While decline in performance due to speech recognition errors can be expected, there is no
guarantee that a speech-aware model yields the same results if speakers vary in age, gender or
dialect, even if the transcription is correct.
A factoid question
that results in an incorrect answer for speakers of a specific demographic group
can be overlooked when summary metrics are analyzed,
although it is an indicator of discriminatory behavior and shows that
the capabilities of the model are not fully retained when accessed via speech input.
The goal of the benchmark presented in this paper is to quantify such behavior.
At first, this task may seem obvious. One could take an ordinary LLM benchmark
such as MMLU \cite{hendrycks2021mmlu}, record the collected samples and pass the results to the audio input of a speech-aware LLM.
However, this approach leads to a gross oversimplification that results in inconclusive outcomes of evaluation.
First, not all textual tasks can be plausibly spoken.
Many popular textual benchmarks are based on exam questions \cite{clark2018arc,hendrycks2021mmlu,hendrycks2021math}.
It is very unlikely that such questions,
preceded by few-shot examples and followed by a list of possible answers,
will be read by the user of a voice interface in its original form.
Second, a comprehensive evaluation of a task in an audio domain requires a diverse group of speakers.
Finally,
raw accuracy is a poor indicator of fair behavior across
different categories of speakers.
A more detailed evaluation is required to verify whether the operation
of the model does not lead to unfair or non-robust behavior.
In order to address the aforementioned problems, we propose the
Cross-modal Capabilities Conservation Test (C3T), a new benchmark
for testing if the language understanding capabilities available in textual LLMs are
present to the same extent in their speech-aware counterparts.
\footnote{The benchmark is available for download at \url{https://github.com/SamsungLabs/C3T}.}

\section{Related Work}

While there are elaborate benchmarks for the vision modality of multimodal
language models, the audio modality remains relatively under-explored.
ASR-GLUE~\cite{feng22asr}, which is a speech-aware variant of the GLUE benchmark, 
aims at
evaluating the impact of speech recognition errors on the performance of natural language understanding models.
Although GLUE tasks can be utilized for
preliminary evaluation, it has to be noted that
the capabilities of textual LLMs are now tested
with more elaborate benchmarks \cite{harness,liang2023holistic}.
Furthermore, the human recordings in ASR-GLUE come from 6 speakers, limiting its suitability for
testing the fairness of the model across speakers' characteristics.
AudioBench \cite{wang-etal-2025-audiobench} is a benchmark for foundational audio models (AudioLLMs)
that aims at evaluating speech understanding, audio scene understanding and audio classification.
The evaluation procedure heavily relies on the use of the LLM-as-a-judge paradigm.
The benchmark reports the performance of AudioLLMs in gender and accent classification,
but it does not track the fairness of the models in a systematic way.
The robustness of models across modalities is not assessed.
AIR-Bench \cite{yang-etal-2024-air} evaluates AudioLLMs
on tasks that approach speech, sound and music-related problems in the form of single-choice and open-ended questions.
The benchmark depends on the use of GPT-4 as a judge to determine the accuracy of models.
Neither fairness among speakers nor robustness across modalities is analyzed.

The official results of the audio modality tests
reported by the commercial suppliers of multimodal LLMs are rather limited.
For GPT-4o, only automatic speech recognition performance and audio translation performance are
reported\footnote{\url{https://openai.com/index/hello-gpt-4o}}, with WER and BLEU used as evaluation
metrics and CoVoST-2 \cite{wang2020covost2} indicated as the dataset used to evaluate audio translation.
The same holds for Gemini 1.5 \cite{geminiteam2024gemini}, which, besides internal datasets, was
evaluated on MLS, FLEURS and CoVoST-2.
The most prominent open-weight multimodal model, Llama 3.1, follows this approach \cite{dubey2024llama31} with speech
recognition and audio translation results reported on MLS, LibriSpeech, VoxPopuli, FLEURS and CoVoST-2.

Fairness analysis is a broad topic with many competing definitions of this term that, in some cases,
are even mutually exclusive \cite[p. 64]{barocas2023fairness}.
The definition of fairness that we adopt in this work resembles the operationalization
of \emph{counterfactual fairness} of \cite{kusner17fairness}
employed by \cite{liang2023holistic} to study the performance of unimodal, textual LLMs.
\cite{kusner17fairness} propose to adopt the definition of fairness that
``captures the intuition that a decision is fair towards an individual if it is the same in
(a) the actual world and (b) a counterfactual world where the individual belonged
to a different demographic group'' \cite[p.1]{kusner17fairness}.
\cite{liang2023holistic} operationalize this notion by aggregating worst-case outcomes determined
for the set of instances derived from the task under study
by substituting lexical terms that indicate gender, dialect or name of a person.
In our formulation, demographic groups are modeled by voices which exhibit different
characteristics, task instances are obtained in the process of voice cloning and outcomes
are binary.

\section{Benchmark Design}

\begin{figure}[htbp]
  \includegraphics[width=\columnwidth]{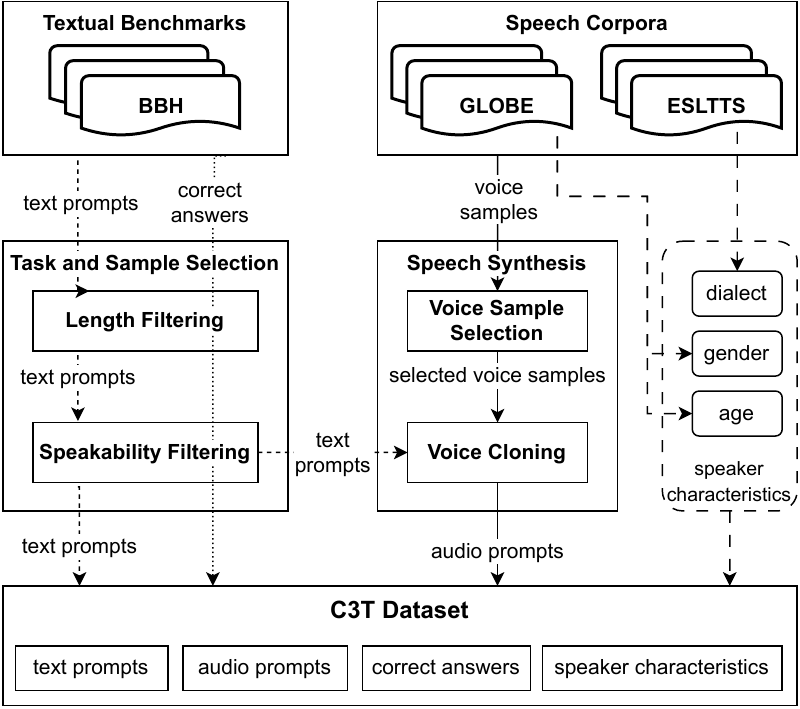}
  \caption{Dataset development}
  \label{fig:development}
\end{figure}

We build our benchmark on the basis of the tasks that are employed to evaluate textual LLMs (Fig.~\ref{fig:development}).
This approach enables us to check if speech-aware models retain the capabilities of textual LLMs and to verify if the behavior of the model under study is robust across modalities.
Tasks from textual benchmarks require special attention when they are utilized to evaluate speech-aware systems as many are not suitable for voice interface without further processing. Therefore, we implemented filtering procedures that enabled us to select tasks that can be plausibly communicated via speech.

We designed the benchmark to be composed of tasks
that have a single, ground truth correct answer that
can be determined by string comparison.
It was a deliberate decision
that eliminated the necessity to use another LLM as a judge,
as this paradigm is still not reliable \cite{white2024livebench,bavaresco-etal-2025-llms} and could
potentially interfere with our baseline model as LLMs tend to prefer the output of their own \cite{panickssery2024llmfavor}.
The answer generated by the model is considered to be correct if it includes the target answer and
does not include other options (if available).

To study the preservation of language understanding capabilities of a speech-aware model,
each textual task has to be executed independently for each category of speakers.
Considering a wide range of tasks to be analyzed and a broad spectrum of
speakers to be investigated,
recording the performance of each speaker in each task would be a massive, infeasible effort.
Therefore, to streamline the evaluation process, we developed a voice cloning text-to-speech model
to synthesize speakers' performances in the tasks under study.
Using voice cloning not only allowed us to design a cost-effective evaluation procedure,
but, as in the case of automatic text sample selection, this makes our benchmark
forward-compatible with tasks that may arise in the future.

As characteristics of speakers that manifest in the voice
can affect the outcome of the evaluated model,
the benchmark for speech-aware models cannot rely on a single coarse-grained measure of the model's
performance such as accuracy.
Thus, we introduced a set of metrics that monitor the performance of the models across
different categories of speakers built on the notion of fairness.
To assess the preservation of language understanding capabilities across speech and text modalities
in a manner that takes into consideration speakers' characteristics, we include
speaker-aware measures based on the idea of cross-modal robustness \cite{kubis-etal-2023-back}.

\section{Tasks}
\label{sec:tasks}

LLMs are being evaluated in a wide range of tasks. Unfortunately, not all of them are plausible use
cases for voice communication. For instance, while several benchmarks incorporate
mathematical tasks \cite{cobbe2021gpqa, hendrycks2021math}, it is not reasonable to assume that the user will dictate the whole, page-long algebraic question from a high-school exam that encompasses complex mathematical equations in order to obtain the solution. Conversely, asking a factual question is an example of a task commonly supported by voice interfaces. Thus, to make the benchmark realistic, we decided to review individual tasks and assess which ones are suitable for audio modality evaluation.
We started with tasks from the Open LLM Leaderboard v2
\cite{open-llm-leaderboard-v2} benchmark because it is a contemporary,
comprehensive and popular benchmark used to compare large language models.
We conducted manual and automatic analysis of the
tasks' suitability to be read aloud. Tasks that were either too long or contained
a prohibitively high ratio of mathematical expressions and/or non-alphanumeric symbols were
eliminated.
Out of a total of 154\,536 samples initially being considered, only 2\,470 text prompts appeared to
be suitable for being spoken.
The samples being preserved
originated from the eleven tasks of BIG-Bench-Hard, a subset of Open LLM Leaderboard v2, that
includes a
very diverse selection of the tasks from the BigBench dataset \cite{suzgun-etal-2023-challenging}.

\section{Metrics}
\label{sec:metrics}

Although we designed the dataset so that the correctness of the model's answer can be verified by
the exact match to the reference value, relying on exact match accuracy as a measure of cross-modal
capabilities transfer would not be a good choice.
Characteristics of speakers that manifest in the voice, but are
lost in transcription, can potentially affect the outcome of the model
in a way that is not possible in a pipeline system that combines Automatic Speech Recognition with a textual LLM model.
While a decline in performance due to speech recognition errors can be expected, there is no
guarantee that
a speech-aware model will act in the same way for speakers that differ in age, gender or dialect,
even if the transcription is correct.
Therefore, our benchmark tracks the fairness of the model.
We consider a task
to be fair for a set of speakers if each instance of the task
returns the same answer regardless of the speaker.
The \emph{overall fairness} of the model in this setting is given by the
following formula:
\begin{equation}
  \frac{\rm\#\ fair\ tasks}{\rm\#\ tasks}
\end{equation}

We also track \emph{conditional fairness}, which is restricted to selected characteristics of
speakers such as age, gender or dialect.
For this purpose, we
assume that the model fulfills the task in a fair manner with respect to characteristic $f$ if
it returns the same answer for each instance of the task regardless of the value of $f$ for the speaker.
Then, we aggregate the results over all the tasks
that do not have speaker-dependent answers:
\begin{equation}
  \frac{\rm\#\ tasks\ fair\ for\ \it f}{\rm\#\ tasks}
\end{equation}

On top of fairness, we define
the notion of cross-modal robustness,
which requires the model to be fair and yield the
same result for speech and text inputs.
We assume the model to be robust across textual and auditory modalities if the model is fair for
the given task and the returned answer matches the result obtained for the text prompt.
\begin{equation}
  \label{eqn:robustness}
  \frac{\rm\#\ fair\ tasks\ that\ match\ text\ results}{\rm\#\ tasks}
\end{equation}

For the purpose of comparison,
we also report textual exact match accuracy for \emph{tasks} that consist of
text prompts paired with
the expected answers and spoken exact match accuracy
for \emph{task instances} that consist of speech samples paired with the
expected answers.

\begin{table*}[!htb]
  \caption{Fairness, robustness and exact match accuracy of the models under study}
  \label{tab:results}
  \centering
  \begin{tabular}{lcrrrrrrr}
    \hline
  \multirow{2}{*}{\textbf{Model}} & \multirow{2}{*}{\textbf{Constrains}} &
    \multicolumn{4}{c}{\textbf{Fairness}} &
    \multirow{2}{*}{\textbf{Robustness}} &
    \multicolumn{2}{c}{\textbf{Exact Match}} \\
    \cline{3-6} \cline{8-9}
    & & \textbf{Overall} & \textbf{Age} & \textbf{Gender} &
    \textbf{Dialect} & & \textbf{Textual} & \textbf{Spoken}\\
    \hline
    Audio Flamingo 3 & none & 0.8287 & 0.8567 & 0.8583 & 0.8522 & 0.6988 & 0.2117 & 0.0766 \\
    Audio Flamingo 3 & c.a. & 0.0024 & 0.0053 & 0.0053 & 0.0082 & 0.0024 & 1.0000 & 0.1717 \\
    \hline
    Qwen2-Audio & none & 0.6510 & 0.7757 & 0.7794 & 0.6907 & 0.6008 & 0.1336 & 0.0634 \\
    Qwen2-Audio & c.a. & 0.0092 & 0.0057 & 0.0057 & 0.0280 & 0.0092 & 1.0000 & 0.3522 \\
    \hline
    Ultravox & none & 0.4275 & 0.5117 & 0.5206 & 0.4644 & 0.4231 & 0.3377 & 0.2215 \\
    Ultravox & c.a. & 0.0160 & 0.0206 & 0.0206 & 0.0427 & 0.0160 & 1.0000 & 0.5818 \\
    \hline
    Voxtral & none & 0.4806 & 0.5964 & 0.6077 & 0.5247 & 0.4773 & 0.1457 & 0.1029 \\
    Voxtral & c.a. & 0.0000 & 0.0016 & 0.0016 & 0.0034 & 0.0000 & 1.0000 & 0.5194 \\
    \hline
  \end{tabular}
\end{table*}

\section{Speech Synthesis}

The ability to evaluate fairness and robustness of a speech-aware model in a systematic way
distinguishes our benchmark from previous approaches.
\cite{liang2023holistic}
addressed the problem of measuring fairness of textual LLMs by taking into
consideration the race, gender and dialect of the speaker.
The evaluation of fairness in the case of speech-aware models requires testing each task on a
diverse group of speakers as demographic characteristics manifest easily in voice.
However, with a large number of tasks to study and speakers' characteristics to consider,
recording each task for each speaker is infeasible due to the high cost.
Therefore, we developed a voice cloning model that enabled us to analyze the performance of
speech-aware models for a wide range of speakers in the tasks that were not uttered by them.
The model is designed after Voicebox \cite{voicebox},
a~text-to-speech model that demonstrates zero-shot capabilities in
reconstructing audio segments based on textual inputs and speech prompts. This model operates
on the Conditional Flow Matching (CFM) \cite{flowmatching}, which utilizes the process of
transitioning from Gaussian noise to a~targeted audio distribution. This transition is
facilitated through the identification of the vector's directional flow within the audio space.
The model's architecture is an adaptation of the transformer \cite{transformer},
incorporating significant modifications compared to its original implementation~\cite{voicebox}. Notably, it diverges
from the use of ALiBi self-attention bias \cite{alibi} and instead integrates
the rotary positional embedding \cite{roformer} over the convolutional positional embedding.

\section{Experiments}
\label{sec:experiments}

For the purpose of illustrating the operation of C3T, we evaluate
Audio Flamingo~3 \cite{goel2025audioflamingo3advancing},
Qwen2-Audio \cite{chu2023qwenaudioadvancinguniversalaudio}, Ultravox
\cite{ultravox2024} and Voxtral Mini \cite{liu2025voxtral} models.
We report overall fairness and robustness of the models under study along with
conditional fairness scores determined with respect to accent, age and gender.
To get a more detailed view, we also present the results constrained to the instances of the tasks
for which at least one
speaker got the correct answer ($c.a.$ constraint in Table~\ref{tab:results}).
For calculation of fairness with respect to accent, we used the ESLTTS \cite{esltts} dataset for
voice cloning, as it contains a variety of accents and speakers. For each text prompt, a speaker
was drawn for each accent, and one sample was then randomly chosen to serve as input for the voice
cloning model.
As a result, for each instance of each task, 29 different
audio prompts were synthesized, each representing a different accent.
In the case of characteristics of gender and age, as a foundation for the voice cloning, the
GLOBE \cite{globe} dataset was used. For each text prompt, for each age, one male and one female
speaker were drawn, and then, for each gender, one audio prompt for the voice cloning model was randomly chosen.
The pool from which the samples were drawn was limited to accents that included at least one
sample for each age range, i.e. Australia, England and Scotland. As there were seven age
ranges and two genders, this approach resulted in synthesis of 14 different audio prompts
for each instance of each task.

The results presented in Table~\ref{tab:results} show that
the transition from text prompts to speech samples leads to a drop in exact match accuracy
ranging from $4\%$ to $13\%$, with Ultravox outperforming the other models in both modalities.
Although the drop in accuracy is significant, it does not reveal if different groups of speakers are
affected in the same way.
Looking at the overall fairness of the models, one
can observe that Audio Flamingo 3 achieves the best result with a score of $0.8287$.
However, if we restrict our attention to tasks for which the model provided a correct answer
to at least one speaker, then the score of $0.0160$ achieved by Ultravox
is the best one.
This means that in more than $98\%$ of tasks that the models can potentially solve via uttered
command some speakers were provided an incorrect answer.
Conditional fairness scores determined for age, gender and dialect
exhibit the same pattern.
More than $51\%$ of tasks are answered in the same way regardless of age, more
than $52\%$ regardless of gender and more than $46\%$ regardless of accent, while no more than
$5\%$ of tasks yield the same result
for any speaker characteristic if the evaluation set is constrained to
tasks that are solved by the model for at least one speaker.
By definition, robustness is limited from the top by overall fairness.
The juxtaposition of fairness and robustness scores
reveals a $13\%$ performance drop in the case of top-performing model (Audio Flamingo~3), indicating that
the model's behavior can be inconsistent across modalities even
when the task is successfully completed for all considered speakers.

\section{Conclusion}

We developed a new benchmark for assessing the performance of speech-aware large language models.
The proposed benchmark relies on automatic extraction, selection and transformation procedures
to determine textual language understanding tasks that can be plausibly passed to the audio input
of a speech-aware LLM and a custom voice cloning model to study the performance of LLMs
across various groups of speakers.
In contrast to previous benchmarks proposed in the literature, which focus on
foundation
properties of speech-aware models such as the ability to recognize speech or classify acoustic
scenes, our benchmark was designed to test the preservation of language understanding capabilities in
the model. Instead of concentrating on raw accuracy reported in other benchmarks, we obtained a more
fine-grained view of the models under study by
investigating fairness across different categories of speakers and robustness
across text and speech modalities.
Our experimental results
confirm that
the model's behavior across
modalities can be inconsistent, even when it answers tasks via speech input in a fair way.

\bibliographystyle{IEEEbib-abbrev}
\bibliography{custom}

\end{document}